\documentclass[10pt,twocolumn,letterpaper]{article}

\usepackage[pagenumbers]{cvpr} 

\usepackage[dvipsnames]{xcolor}

\usepackage[accsupp]{axessibility}
\definecolor{cvprblue}{rgb}{0.21,0.49,0.74}
\usepackage[pagebackref,breaklinks,colorlinks,citecolor=cvprblue]{hyperref}
\usepackage{url}            
\usepackage{booktabs}       
\usepackage{amsfonts}       
\usepackage{nicefrac}       
\usepackage{microtype}      
\usepackage{xcolor}         

\usepackage{cite}
\usepackage{graphicx}
\usepackage{amsmath}
\usepackage{amssymb}
\usepackage{epsfig}
\usepackage{algorithm}  
\usepackage{algpseudocode}
\usepackage{pifont}
\usepackage{multirow,multicol, makecell}
\usepackage{bbm}
\usepackage{mathtools}
\usepackage{diagbox}
\usepackage{enumitem}
\usepackage{soul}

\newcommand{\gray}[1]{{\textcolor[RGB]{180,180,180}{#1}}}

\def\paperID{4690} 
\def\confName{CVPR}
\def\confYear{2024}

\title{UniPT: Universal Parallel Tuning for Transfer Learning with \\ Efficient Parameter and Memory}

\author{%
  Haiwen Diao$^1$, 
  Bo Wan$^2$, 
  Ying Zhang$^3$, 
  Xu Jia$^1$, 
  Huchuan Lu\thanks{Corresponding author. Work was done when Haiwen visited HKUST.}\: $^1$,
  Long Chen$^4$
  \\
  $^1$Dalian University of Technology \; 
  $^2$KU Leuven \; 
  $^3$Tencent WeChat \;
  $^4$HKUST \\
  {\tt\small diaohw@mail.dlut.edu.cn; 
          bwan@esat.kuleuven.be;
          yinggzhang@tencent.com;
          } \\
  {\tt\small xjia@dlut.edu.cn; \;\,\,
          lhchuan@dlut.edu.cn; \;\,\,
          longchen@ust.hk} \\
  }

\begin{document}
\maketitle

\begin{abstract}
Parameter-efficient transfer learning (PETL), i.e., fine-tuning a small portion of parameters, is an effective strategy for adapting pre-trained models to downstream domains. To further reduce the memory demand, recent PETL works focus on the more valuable memory-efficient characteristic. In this paper, we argue that the scalability, adaptability, and generalizability of state-of-the-art methods are hindered by structural dependency and pertinency on specific pre-trained backbones. To this end, we propose a new memory-efficient PETL strategy, Universal Parallel Tuning (\textbf{UniPT}), to mitigate these weaknesses. Specifically, we facilitate the transfer process via a lightweight and learnable parallel network, which consists of: 1) A parallel interaction module that decouples the sequential connections and processes the intermediate activations detachedly from the pre-trained network. 2) A confidence aggregation module that learns optimal strategies adaptively for integrating cross-layer features. We evaluate UniPT with different backbones (e.g., T5~\cite{TransF:T5}, VSE$\infty$~\cite{ITM:GPO}, CLIP4Clip~\cite{VLP:CLIP4Clip}, Clip-ViL~\cite{VLP:CLIP-ViL}, and MDETR~\cite{VLP:MDETR}) on various vision-and-language and pure NLP tasks. Extensive ablations on \textbf{18} datasets have validated that UniPT can not only dramatically reduce memory consumption and outperform the best competitor, but also achieve competitive performance over other plain PETL methods with lower training memory overhead. Our code is publicly available at: \href{https://github.com/Paranioar/UniPT}{https://github.com/Paranioar/UniPT}.
\end{abstract}

\section{Introduction}

Large-scale deep neural networks~\cite{TransF:GPT,TransF:BERT,TransF:ViT,VLP:CLIP} trained on massive data have been successfully investigated in various vision and multimodal learning tasks~\cite{Datasets:MSCOCO,Datasets:MSRVTT,Datasets:VQAv2,Datasets:GQA,Datasets:NLVR,Datasets:VLN}. The most prevalent and straightforward strategy for transferring the knowledge from pre-trained models to downstream tasks is fully fine-tuning~\cite{ITM:GPO,VLP:MDETR,VLP:CLIP-ViL,VLP:CLIP4Clip}. However, fine-tuning the entire network is prohibitively expensive, especially given a large model with millions of parameters. Meanwhile, it easily suffers from the over-fitting problem with a relatively ``small'' downstream dataset. 
To address it, there is an increasing interest in \emph{parameter-efficient transfer learning} (PETL)~\cite{TL:Adapter-NMT,TL:AdapterFusion,TL:LoRA,TL:Prefix-Tuning}, which facilitates domain adaptation by adjusting or inserting a few modules.

\begin{figure*}[!t]
    \centering \includegraphics[height=0.37\linewidth,trim= 0 150 0 0,clip]{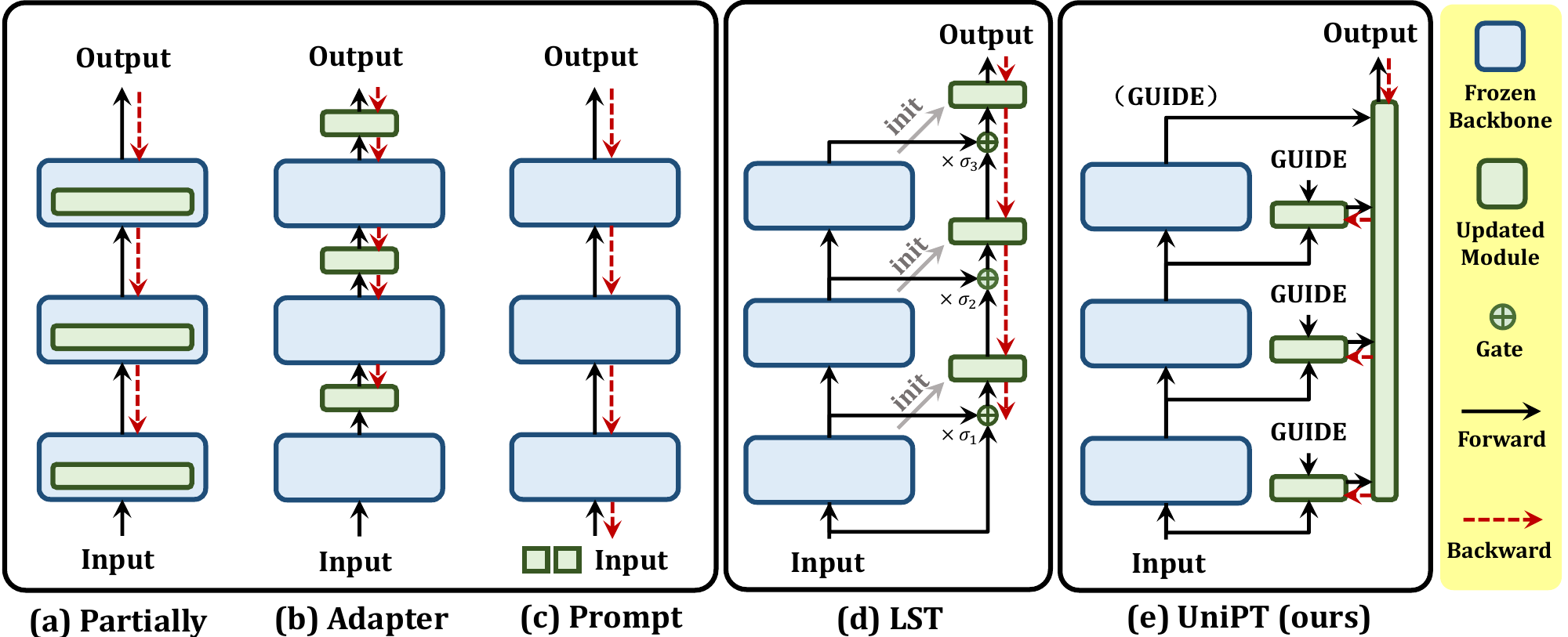}
    \vspace{-0.5em}
    \caption{Overview of recent PETL methods including \emph{Partially Tuning}, \emph{Adapter Tuning}, \emph{Prompt Tuning}, and \emph{Ladder Side Tuning (LST)}.}
    \label{fig:motivation}
\end{figure*}

Currently, mainstream state-of-the-art PETL approaches can be coarsely grouped into three categories: 
(a) \textbf{Partially Tuning}~\cite{TL:BitFit,TL:Layernorm-Tuning}: It only updates a few task-specific parameters and freezes most original parameters, such as modifies the bias items~\cite{TL:BitFit,TL:Tiny-TL} or the normalization layers~\cite{TL:Layernorm-Tuning}.
(b) \textbf{Adapter Tuning}~\cite{TL:Adapter-BERT,TL:LoRA,TL:Compacter,TL:CLIP-Adapter}: It usually inserts a new bottleneck-shaped module after each backbone layer, which is the only part that needs to be updated. Typically, the new module consists of a linear down-projection, a non-linearity activation, and a linear up-projection.
(c) \textbf{Prompt Tuning}~\cite{TL:Prefix-Tuning,TL:PPT,TL:VPT,TL:CoOp}: 
It first integrates a fixed number of learnable vectors as additional input tokens (\ie, prompts). Then, it only learns the prompts and freezes all the raw parameters of the pre-trained network during the fine-tuning stage.
Although all the above PETL methods can dramatically reduce the trainable parameters and storage constraints, their memory consumption remains costly during the training stage. As shown in~\Cref{fig:motivation}(a-c), the backward gradients still need to go through (nearly the entire) foundation models. This memory-extensive characteristic severely limits their applications in resource-constrained scenarios.

Consequently, a few recent works~\cite{TL:Side-Tuning,TL:Tiny-TL,TL:Y-Tuning,TL:LST,TL:InCA,TL:VQT} emphasize the requirements of both \emph{parameter} and \emph{memory} efficiency during the training process. In particular, the most successful method that achieves a good balance between performance and efficiency is Ladder Side Tuning (\textbf{LST})~\cite{TL:LST}.
As shown in~\Cref{fig:motivation}(d), it first constructs a lightweight side network by keeping the same structure as the pre-trained network but reducing the dimension of each original layer by a predefined reduction factor. Then, it learns a static gated ladder connection to combine the pair-wise features at each layer between the side and pre-trained network.

Nevertheless, we argue that these designs have several potential drawbacks: 
1) \emph{Scalability}:
The model complexity of the side network goes linearly proportional to the original pre-trained network, making its efficiency susceptible to the original architecture, \ie, the larger the pre-trained network, the less efficient the side network.
2) \emph{Adaptability}: 
The static gate mechanism simply sums up the outputs of each pre-trained and its corresponding side layer. It overlooks the latent semantic misalignments of pair-wise features between them, and also neglects to dynamically equip different samples with the most suitable aggregation strategy.
3) \emph{Generalizability}:
Most of the above strategies are primarily suitable for the Transformer-family. For instance, in other prevalent neural networks (\eg, CNNs), the ladder gate connection in LST cannot directly handle the discrepancy of spatial-wise and channel-wise dimensions between cross-layer features. Therefore, how to extend these PETL methods to various architectures is still under-explored.

Based on such considerations, in this paper, we propose a new memory-efficient PETL strategy, dubbed Universal Parallel Tuning (\textbf{UniPT}).
As shown in~\Cref{fig:motivation}(e), we facilitate the transfer process via a lightweight learnable parallel network, whose structures are independent of the backbone (\emph{Scalablity}) and could work across various architectures, such as Transformers, CNNs and Encoder-Decoder structures (\emph{Generalizability}). Specifically,
our UniPT consists of two modules: 1) A parallel interaction module decouples the inherently sequential connections across layers and handles each layer's intermediate detachedly. It treats the feature outputs across layers equally and highlights more discriminative representations inside each layer.
2) A confidence aggregation module learns adaptively optimal strategies for integrating cross-layer features based on different input embeddings and model structures. It requires no manual tuning while staying effective and efficient (\emph{Adaptability}).

To fully evaluate the generalization capability, we have made comprehensive investigations on both challenging cross-modal and widely-used uni-modal domains by typically involving more diverse network architectures than previous works. 
Extensive results on vision-and-language tasks (\ie, \emph{image-text retrieval}~\cite{Datasets:MSCOCO,Datasets:Flickr30k}, \emph{video-text retrieval}~\cite{Datasets:MSRVTT,Datasets:MSVD}, \emph{visual question answering}~\cite{Datasets:VQAv2}, \emph{compositional question answering}~\cite{Datasets:GQA}, and \emph{visual grounding}~\cite{Datasets:REFCOCO,Datasets:REFCOCOG}) and one GLUE benchmark have validated the effectiveness of our UniPT, \ie, it achieves the best balance and trade-off between performance and parameter/memory efficiency.

\section{Related Work}

\noindent\textbf{Parameter-Efficient Transfer Learning (PETL).}
Recently, large pre-trained models have sprung up in computer vision (CV)~\cite{TransF:ViT,TransF:SwinTransformer,CL:MoCo,CL:DINO}, natural language processing (NLP)~\cite{TransF:BERT,TransF:RoBERTa,TransF:T5}, and vision-and-language (VL) fields~\cite{VLP:CLIP,VLP:ALBEF,VLP:MDETR,VLP:BLIP}. 
To efficiently transfer pre-trained knowledge, PETL methods have drawn much research attention and become a promising direction. 
Earlier PETL studies~\cite{TL:BitFit,TL:Layernorm-Tuning} attempted to update a few pre-trained network parameters during fine-tuning including the bias parameters~\cite{TL:BitFit,TL:Tiny-TL} and layer normalization layers~\cite{TL:Layernorm-Tuning}. However, they are not always applicable, \eg, the pre-trained T5 model~\cite{TransF:T5} does not have any bias items. 
Hence, some works~\cite{TL:FISH-Mask,TL:DiffPruning} tried to control which components inside the large network to be updated or fixed during training by learning sparse binary marks against the pre-trained weights. Nevertheless, the resulting performance is extremely sensitive to the sparsity of the binary mask. 

In contrast, another line of research introduced an extra lightweight learnable subnetwork (Adapter) that is integrated into the original network layers and arouses a surge of interest in NLP~\cite{TL:Adapter-BERT,TL:Adapter-NMT,TL:AdapterFusion,TL:AdaLoRA}, CV~\cite{TL:CLIP-Adapter,TL:AdaptFormer,TL:ViT-Adapter,TL:SSF}, and VL areas~\cite{TL:VL-ADAPTER,TL:UniAdapter,TL:Tem-adapter,TL:VL-PET,TL:VLN-PETL}. 
They typically injected a tiny bottleneck module in parallel or after each multi-head attention and feed-forward layer, and kept the rest of the pre-trained network frozen. 
Particularly, some works~\cite{TL:LoRA,TL:IA3,TL:UnifiedPET} inserted a trainable vector or low-rank matrix into multi-head attention to influence the query and value projection, while other methods~\cite{TL:Compacter,TL:FacT,TL:Hypernetwork} implemented matrix decomposition, low-rank parameterization, and hyper-network generation for the adapter weights respectively to further reduce the number of parameters that need to be trained.
Concurrently, another popular technology namely Prompt Tuning~\cite{TL:Prefix-Tuning,TL:PEPT,TL:PPT} in NLP validated that simply prepending several learnable tokens into the input sequence of each Transformer layer can also achieve competitive performance during fine-tuning, which further triggers a series of follow-up explorations in CV~\cite{TL:VPT,TL:CoCoOp,TL:DenseCLIP,TL:MaPLe,TL:PLOT} and VL fields~\cite{TL:Frozen,TL:Efficient-Prompt,TL:VoP,TL:ScienceQA,TL:MM-CoT}.
The most critical issue is that the optimal prompt module requiring elaborate manual tuning and designs is extremely challenging for specific downstream datasets.
Although the aforementioned PETL methods significantly decreased the trainable parameters and produced remarkable results on the downstream task, they suffer from expensive computational memory consumption, making them inapplicable to resource-constrained scenarios during training.

\noindent\textbf{Memory-Efficient Transfer Learning.}
Current PETL methods investigate the ways to achieve competitive performance with as few trainable parameters as possible. Nevertheless, the training memory is dominated by activations, not parameters, which means that parameter efficiency is not equivalent to memory efficiency. Hence, some approaches~\cite{TL:ULMFiT,VLP:CLIP,TransF:T5,CL:SimCLR} fixed the pre-trained backbone and only updated the last layers for domain transfer, which, though memory-efficient, have limited model capacity and lag far behind the results by fully fine-tuning. Hence, Side-Tuning~\cite{TL:Side-Tuning} adopted a cheap side network, whose outputs are combined with the backbone outputs with a curriculum schedule for continuous task adaptation. Besides, Y-tuning~\cite{TL:Y-Tuning} exhausted all possible labels and fed their dense representations via a side feature integration for the final label selection. To address intractable answer collection and focus more on memory reduction, LST~\cite{TL:LST} proposed a small and separate network that receives intermediate activations from backbone networks via ladder gated connections and makes predictions. As mentioned before, our UniPT outperforms their solutions in terms of Scalability, Adaptability, and Generalizability, which displays more powerful capability and broad applicability over various model architectures in multiple VL tasks.
Actually, there are some popular and generalized  strategies~\cite{transF:Reformer,TransF:Rev-ViT} to alleviate the training memory footprint. Some of them avoid saving all the intermediate activations and re-compute discarded ones during backward via reconstruction from the backward layers~\cite{CNN:RevNet} or gradient checkpoint operation~\cite{Training:Sublinear}, while others~\cite{Training:8-bit} developed new reduced-precision floating-point formats to decrease the bitwidth of training activations.
Note that our UniPT is orthogonal to these techniques and can be combined to further pursue a higher level of memory efficiency.

\begin{figure*}[!t]
    \centering \includegraphics[height=0.40\linewidth,trim= 0 320 455 5,clip]{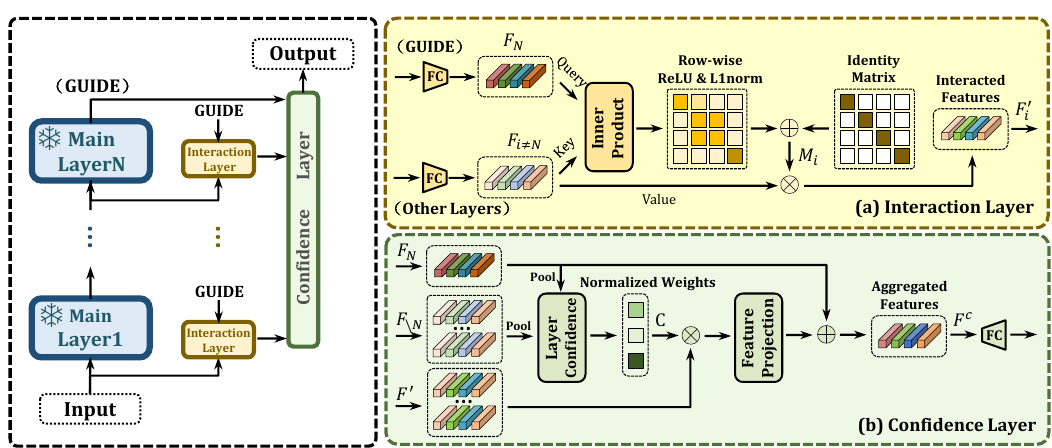}
    \vspace{-0.5em}
    \caption{Overview of the Framework with (a) parallel interaction and (b) confidence aggregation layers. The former attempts to extract more discriminative features at each layer independently guided by the relatively most powerful output features, while the latter learns a dynamic and optimal combination strategy over the blended features at each layer for the ultimate domain adaptation.}
    \label{fig:framework}
\end{figure*}

\section{Methodology}
We propose a novel Universal Parallel Tuning (UniPT) for memory-efficient transfer learning.
In~\Cref{subsec:unipt}, we explain the detailed architecture and implementation of UniPT including a parallel interaction layer and a confidence aggregation layer. Then, we demonstrate its applications to CNN, Transformer, and Encoder-Decoder network in~\Cref{subsec:application}.

\subsection{Universal Parallel Tuning}
\label{subsec:unipt}

As illustrated in~\Cref{fig:framework}, we construct a tiny and parallel network on top of the pre-trained network that could transform all the intermediate layer features into the final representations for new domains. It does not require costly backward gradients through the large pre-trained backbone and vastly lessens memory overhead during the training process.

\textbf{Parallel Interaction Layer.}
We build parallel interaction layers that are lightweight and independent of the pre-trained network. 
In~\Cref{fig:framework}(a), they take each layer's intermediate representation from the pre-trained backbone as input and handle the feature extraction of each layer detachedly.

Existing memory-efficient PETL methods~\cite{TL:Side-Tuning,TL:LST,TL:Y-Tuning} implement self-attention inside scaled-down side blocks and incorporate hierarchical features in a bottom-up manner. 
However, direct lateral connection might neglect the potential semantic inconsistency between side and pre-trained backbones. More significantly, the weakened interaction would dilute the discriminative representations within each layer necessary for ultimate adaptation.
Inspired by this, we conduct a top-down interaction process, leveraging ultimate outputs from the pre-trained network as interaction guidance.
This choice stems from the superior representation and transfer capability exhibited by the original last layer outputs in new domains.
By utilizing them as uniform guidance for other layers before hierarchical aggregation, it could ensure semantic coherence across different layers and capture the essential features inside each layer independently, thereby in turn enhancing the ultimate representations accordingly.

Concretely, we first map all $K$ hidden features of all $N$ layers to a unified dimension $d=D/r$ via a reduction factor $r$, where $D$ is the dimension of the original backbone output. 
We empirically observe a significant disparity among feature norms across different layers within various backbones. Computation in the standard attention layer would suffer numerical issues and unstable gradients during backpropagation.
Hence, after obtaining all the layer features $\boldsymbol{F}=\{\boldsymbol{F}_{i}\in\mathbbm{R}^{K\times d} \, | \, i\in\{0, 1, ..., N\}\}$, we first compute the inner product matrices between last layer $\boldsymbol{F}_{N}$ (Query) and other layers $\boldsymbol{F}_{\tiny\backslash N}=\{\boldsymbol{F}_{i}\in\mathbbm{R}^{K\times d} \, | \, i\in\{0, 1, ..., N-1\}\}$ (Key). Then, we adopt ReLU activation ($\sigma$) and L1 normalization (\texttt{L1Norm}) to eliminate all the negative connections and generate the normalized attention weights between a Query and all Key features. 
Considering that these weights occasionally are all zeros, we add an extra identity matrix bias (\ie residual connection with original layer features) to obtain the final attention weights $\boldsymbol{M} = \{\boldsymbol{M}_{i}\in\mathbbm{R}^{K \times K} \, | \, i\in\{0, 1, ..., N-1\}\}$. Lastly, the blended features $\boldsymbol{F}^{\prime}=\{\boldsymbol{F}_{i}^{\prime}\in\mathbbm{R}^{K\times d} \, | \, i\in\{0, 1, ..., N-1\}\}$ are calculated as follows:
\begin{equation}
\label{eq:inner_product}
\boldsymbol{M}_{i} = \texttt{L1Norm}_{\boldsymbol{F}_{i}} \sigma(\boldsymbol{F}_{N}\boldsymbol{F}_{i}^{\top}) + \boldsymbol{I} \, , 
\quad \boldsymbol{F}_{i}^{\prime} = \boldsymbol{M}_{i}\boldsymbol{F}_{i} \, ,
\end{equation}
where $\boldsymbol{F}_{i}^{\prime}$ denotes $K$ merging features in the $i$-th layer, each of which corresponds the one in $\boldsymbol{F}_{\tiny\backslash N}$.

\textbf{Confidence Aggregation Layer.}
The sequential ladder-gated connection~\cite{TL:LST} as static fusion strategy fails to dynamically adapt to different inputs and suffers from exponential attenuation of earlier layer features. Therefore, we explore an adaptive confidence module that treats the merging features from each layer equally and automatically learns the optimal aggregation strategies for various inputs and model structures. In this way, it can highlight more discriminative features and discard less informative ones from the multi-granularity layers for better domain adaptation. Given the blend features $\boldsymbol{F}^{\prime}$ and $\boldsymbol{F}$, we first obtain the holistic representations of each layer $\boldsymbol{F}^{g}=\{\boldsymbol{F}_{i}^{g}\in\mathbbm{R}^{1\times d} \, | \, i\in\{0, 1, ..., N\}\}$ by averaging all the features $\boldsymbol{F}$. Comparing the relations between last layer $\boldsymbol{F}^{g}_{N}$ and other layers $\boldsymbol{F}^{g}_{\tiny\backslash N}$, we obtain the normalized confidence weights $\boldsymbol{C}$ of the blended features $\boldsymbol{F}^{\prime}$ via a fully-connected (FC) layer and a softmax function. Once all the blended features are merged, they are converted into a shared space as $\boldsymbol{F}_{N}$ via $\texttt{MLP}(\boldsymbol{X})$ = $\sigma(\boldsymbol{X}\boldsymbol{W}_{1})\boldsymbol{W}_{2}$, where $\boldsymbol{W}_{1}\in\mathbbm{R}^{d\times D}, \boldsymbol{W}_{2}\in\mathbbm{R}^{D\times d}$ as follows:
\begin{equation}
\label{eq:confidence}
\begin{split}
\boldsymbol{C} &= \texttt{Softmax}_{\boldsymbol{F}^{g}_{\tiny\backslash N}} ((\boldsymbol{F}_{N}^{g} \odot\boldsymbol{F}^{g}_{{\tiny\backslash} N})\boldsymbol{W}_{3}) \, , \\
\boldsymbol{F}^{c} &= \texttt{MLP}(\sum\nolimits_{i}\boldsymbol{C_{i}}\cdot\boldsymbol{F}_{i}^{\prime}) + \boldsymbol{F}_{N} \, .
\end{split}
\end{equation}
Note that $\boldsymbol{W}_{3}\in\mathbbm{R}^{d\times 1}$ and $\odot$ denotes element-wise multiplication. Eventually, we sum up them and the original output $\boldsymbol{F}_{N}$ as $\boldsymbol{F}^{c}\in\mathbbm{R}^{K\times d}$, which are upsampled by a new FC layer to produce the final outputs. Note that the dimension of the UniPT output is consistent with that of the raw pre-trained backbone for the new task domains.

\begin{figure*}[!t]
    \centering \includegraphics[height=0.33\linewidth,trim= 0 305 270 0,clip]{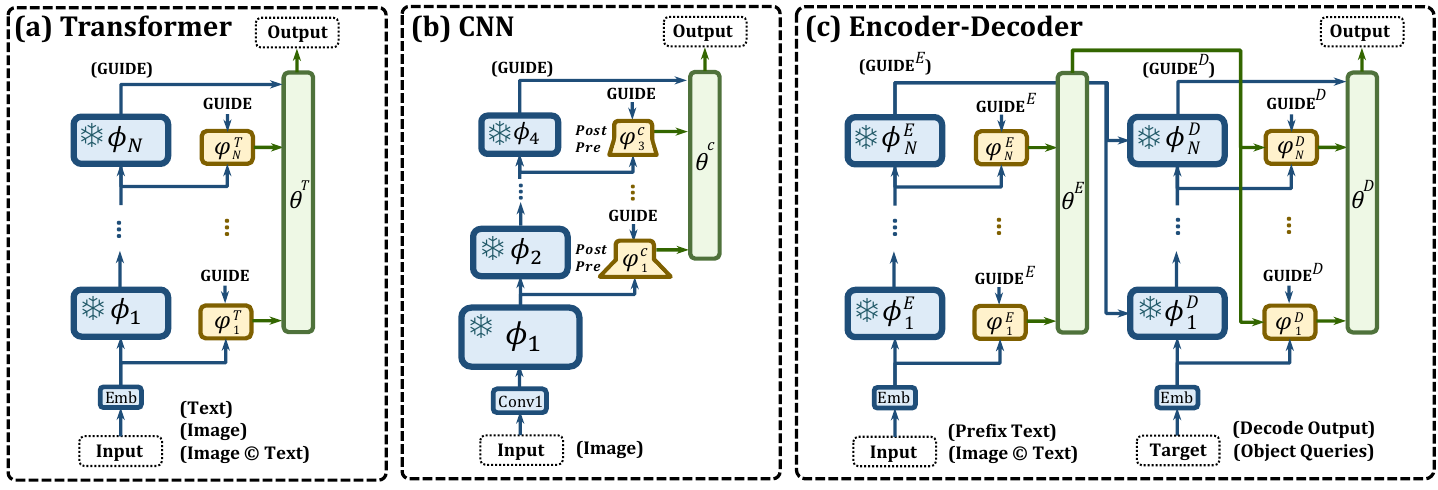}
    \vspace{-0.5em}
    \caption{Overview of the Application with UniPT network ($\varphi,\theta$) over (a) Transformer, (b) CNN, and (c) Encoder-Decoder ($\phi$) architectures.}
    \label{fig:application}
\end{figure*}

\subsection{Diverse Backbone Application}
\label{subsec:application}

As illustrated in~\Cref{fig:application}, our proposed method has broad applications over various frameworks $\phi$, including CNN, Transformer, and Encoder-Decoder architectures. The proposed UniPT consists of two main parts: parallel interaction layer $\varphi$ and confidence aggregation layer $\theta$.

\textbf{Application on Transformer.} In~\Cref{fig:application}(a), we demonstrate that the UniPT can be seamlessly integrated into existing Transformer architectures. Thereinto, for mono-modality encoders like BERT~\cite{TransF:BERT} or ViT~\cite{TransF:ViT} with text or image as input, we extract the word or patch embeddings, all the hidden states, and the original last state, which are then fed to the respective $\varphi_{1:N}^{T}$ layers and integrated into the ultimate feature output by the $\theta^{T}$ layer. For cross-modality correspondence like CLIP-ViL~\cite{VLP:CLIP-ViL}, the image and text features are first mapped into the same dimensions and concatenated into the follow-up cross-Transformer $\phi_{1:N}$ layers. Analogously maintaining the same numbers and dimensions as the pre-trained Transformer output, the final feature representation is enhanced via the parallel $\varphi_{1:N}^{T}$ and $\theta^{T}$ layers.

\textbf{Application on CNN.} For CNN (\eg ResNeXt~\cite{CNN:ResNeXt17}), it processes image inputs with a 2-D convolution kernel where each layer has varied block numbers and structures, as shown in~\Cref{fig:application}(b). More importantly, the spatial sizes and channel dimensions of intermediate feature maps across the layers are different and doubled in value. Therefore for cross-layer feature maps, it is impossible to perform one-to-one gated addition like LST, but it does not interfere with our UniPT owing to the separate processing.
Note that directly employing interaction layer $\varphi_{1:3}^{C}$ on the substantial shallow features would inevitably increase the computational cost. To further improve memory efficiency, we decompose the standard interaction layer as pre-interaction and post-interaction. 
Specifically, we start by dividing the shallow feature map from each layer into multiple non-overlapping chunks, of which the number aligns with the size of the output feature map. 
Given that the representations within each chunk are most relevant to the corresponding positional feature within the last feature map due to the relatively limited receptive field, we thereby implement the interaction process inside a confined region between the last feature and its corresponding shallow chunk, termed as pre-interaction. With the feature maps from Conv2-5 $\phi_{1:4}$ layers, we totally obtain four spatial-reduced feature representations with the same spatial resolution, that then pass the same procedure as the usage in Transformer, denoted as post-interaction.

\textbf{Application on Encoder-Decoder.} As an extension of the conventional Transformer, the pre-trained encoder-decoder model in~\Cref{fig:application}(c) serves as the particular backbone for auto-regressive tasks (\eg, MDETR~\cite{VLP:MDETR} for multi-modal detection and T5~\cite{TransF:T5} for GLUE benchmark). For the Encoder, we implement the Encoder UniPT ($\varphi^{E},\theta^{E}$) with both image and text features in the same way that it does in the Transformer, while for the Decoder, the inputs for the Decoder UniPT ($\varphi^{D},\theta^{D}$) includes an additional output of Encoder UniPT module. Inspired by the original backbone workflow, we also decompose the whole interaction layer $\varphi^{D}$ as pre-interaction and post-interaction as well. Concretely, we first perform the pre-interaction between the Decoder intermediate features and the Encoder UniPT output to enhance the feature representations of each Decoder layer, and then accomplish the post-interaction under the guidance of the original Decoder output for the final domain adaptation.

\addtolength{\tabcolsep}{-2pt}
\begin{table*}[t]\small
    \centering
    \scalebox{0.95}{
    \begin{tabular}{lccccc|ccccc|ccccc}
    \toprule
    \multirowcell{2}{Method} 
    & Params. 
    & Memory 
    &\multicolumn{3}{c|}{BERT} 
    & Params. 
    & Memory 
    &\multicolumn{3}{c|}{ResNeXt-101} 
    & Params. 
    & Memory 
    &\multicolumn{3}{c}{ViT + Text Transf.} \\
    \cmidrule{4-6} \cmidrule{9-11} \cmidrule{14-16}
    &(M) &(G) 
    &I-T &T-I &Rsum 
    &(M) &(G) 
    &I-T &T-I &Rsum
    &(M) &(G) 
    &T-V &V-T &Rsum\\
    \midrule
    \gray{Fully-FT} 
    &\gray{109.5} &\gray{9.9} 
    &\gray{79.7} &\gray{62.1} &\gray{513.5}
    &\gray{90.9} &\gray{21.8 * 8} 
    &\gray{85.6} &\gray{70.2} &\gray{539.0} 
    &\gray{151.3} &\gray{12.2 * 4} 
    &\gray{42.8} &\gray{42.1} &\gray{389.2} 
    \\
    Adapter~\cite{TL:Adapter-BERT}
    &2.6 &8.8 
    &79.1 &60.5 &511.3
    &3.5 &22.1 * 8 
    &66.8 &62.9 &493.3
    &5.2 &10.3 * 4
    &38.3 &39.6 &364.3
    \\
    LoRA~\cite{TL:LoRA}
    &1.1 &8.8 
    &78.8 &59.6 &508.2
    &--&-- &--&--&--
    &1.3 &10.2 * 4
    &38.8 &39.9 &366.8
    \\
    BitFit~\cite{TL:BitFit} 
    &0.9 &8.6 
    &77.3 &57.8 &503.9
    &2.2 &21.0 * 8 
    &83.4 &67.4 &530.6
    &0.1 &10.5 * 4
    &38.1 &40.6 &370.8
    \\    
    Prompt~\cite{TL:Prefix-Tuning}
    &10.7 &9.4 
    &78.7 &59.0 &508.5
    &--&-- &--&--&--
    &0.2 &10.7 * 4
    &36.8 &37.5 &358.8
    \\
    SSF~\cite{TL:SSF}
    &0.2 &8.4
    &80.0 &60.4 &512.8
    &0.1 &20.4 * 8 &83.7&66.8&528.5
    &0.5 &9.8 * 4
    &40.2 &41.8 &376.6
    \\
    FacT~\cite{TL:FacT}
    &0.6 &8.7
    &79.2 &59.3 &508.8
    &--&-- &--&--&--
    &0.8 &10.2 * 4
    &38.7 &39.8 &367.2 
    \\
    AdaLoRA~\cite{TL:AdaLoRA}
    &1.0 &8.8
    &79.8 &60.1 &510.3
    &--&-- &--&--&--
    &1.2 &10.5 * 4
    &39.2 &39.6 &368.5 
    \\
    \midrule
    \gray{Partially$\Downarrow$} 
    &\gray{0.8} &\gray{1.0} 
    &\gray{74.8} &\gray{53.7} &\gray{485.5}
    &\gray{2.1} &\gray{14.9} 
    &\gray{75.2} &\gray{58.2} &\gray{505.8}
    &\gray{0.7} &\gray{1.9 * 4}
    &\gray{36.4} &\gray{37.0} &\gray{353.9}
    \\
    LST~\cite{TL:LST}
    &7.5 &4.6 
    &77.9 &57.3 &501.9
    &2.27 &15.0 
    &82.3 &66.1 &526.7
    &11.2 &8.0 * 4
    &37.0 &37.8 &356.7
    \\    
    \textbf{UniPT}
    &5.9 &3.1 
    &\textbf{80.2} &\textbf{59.8} &\textbf{510.5}
    &6.4 &15.0 
    &\textbf{84.0} &\textbf{67.7} &\textbf{532.1}
    &9.6 &3.4 * 4
    &\textbf{38.9} &\textbf{39.3} &\textbf{361.3}
    \\
    \bottomrule
    \end{tabular}
    }
    \vspace{-0.8em}
    \caption{Comparisons with popular parameter/memory-efficient methods on Flickr30K using \emph{VSE$\infty$} with CNN {\scriptsize(ResNeXt-101)} or single Transformer {\scriptsize(BERT)}, and MSR-VTT using \emph{CLIP4Clip} with dual Transformer encoders {\scriptsize(ViT + Text Transformer)}. 
    We report Recall@1 (R@1) on sentence retrieval (``I-T'', ``V-T''), image retrieval (``T-I''), video retrieval (``T-V''), and ``Rsum'' of R@1,5,10 on bi-directional retrievals.}
    \label{tab:VL_PETL}
    \vspace{-0.5em}
\end{table*}
\addtolength{\tabcolsep}{2pt}

\begin{table*}[t]\small
    \centering
    \resizebox{\textwidth}{!}{
    \begin{tabular}{lcccccccccccc}
    \toprule
    \multirowcell{2}{Method} 
    & Update Params. 
    &\multicolumn{2}{c}{Memory (G)} 
    &\multirowcell{2}{CoLA}
    &\multirowcell{2}{SST-2}
    &\multirowcell{2}{MRPC} 
    &\multirowcell{2}{QQP} 
    &\multirowcell{2}{MNLI}
    &\multirowcell{2}{QNLI}
    &\multirowcell{2}{RTE} 
    &\multirowcell{2}{STS-B}
    &\multirowcell{2}{Avg.}\\ 
    & per Task (\%) &Train &Test \\
    \midrule
    \gray{Fully-FT} 
    & \gray{100} & \gray{17.6} & \gray{0.86}
    & \gray{62.8} & \gray{93.9} & \gray{91.9} & \gray{89.9} & \gray{86.2} & \gray{92.5} & \gray{74.1} & \gray{90.3} & \gray{85.2} \\
    Adapter~\cite{TL:Adapter-BERT} 
    & 1.63 & 13.0 &0.87
    & 64.4 & 94.2 & 88.9 & 88.9 & 86.4 & 93.1 & 75.1 & 91.1 & 85.3 \\
    LoRA~\cite{TL:LoRA} 
    & 1.71 & 12.6 &0.86
    & 63.3 & 94.3 & 90.1 & 89.0 & 86.3 & 93.2 & 75.5 & 90.9 & 85.3 \\
    BitFit~\cite{TL:BitFit} 
    & 0.13 & 10.7 &0.86
    & 61.8 & 94.3 & 91.0 & 88.7 & 85.6 & 93.1 & 67.6 & 90.8 & 84.1 \\
    Prompt~\cite{TL:Prefix-Tuning} 
    & 0.03 & 22.2 &0.87
    & 0 & 90.3 & 74.6 & 88.5 & 82.5 & 92.5 & 59.5 & 90.1 & 72.2 \\
    \midrule
    LST~\cite{TL:LST} 
    &1.74 & 5.5 &0.88
    & 58.1 & 94.1 & 90.4 & 88.8 & \textbf{85.6} & \textbf{93.3} & \textbf{71.9} & \textbf{90.7} & 84.1 \\
    \textbf{UniPT} 
    &1.36 &2.9 &0.86
    &\textbf{62.2} &\textbf{94.2} &\textbf{90.8} &\textbf{88.9} &85.5 &\textbf{93.3} &69.8 &89.7 &\textbf{84.3} \\
    \midrule  
    LST~\cite{TL:LST} (T5-large)
    & 1.23 & 12.2 &2.88
    & 65.3 & 95.7 & 91.6 & \textbf{89.7} & \textbf{88.6} & 94.1 & \textbf{79.9} & \textbf{92.4} & 87.1 \\
    \textbf{UniPT (T5-large)}
    &0.92 &9.1 &2.82
    &\textbf{65.7} &\textbf{95.8} &\textbf{92.0} &\textbf{89.7} &88.2 &\textbf{94.2} &79.6 &92.0 &\textbf{87.2} \\
    \bottomrule
    \end{tabular}
    }
    \vspace{-0.8em}
    \caption{Comparisons with parameter/memory-efficient methods on GLUE benchmark using encoder-decoder \emph{T5}. Following LST, we adopt \emph{T5-base} as baseline and larger \emph{T5-large} for further comparison. We report accuracy for SST-2, MNLI, QNLI and RTE. For CoLA and STS-B, we use Matthew’s Correlation and Pearson-Spearman Correlation respectively. For MRPC and QQP, we average the F1 score and accuracy.}
    \label{tab:GLUE_PETL}
\end{table*}

\section{Experiments}
\label{sec:experiments}

\subsection{Setups}
\label{subsec:setup}

\textbf{Datasets.}
We evaluate our proposed UniPT on VL and NLP tasks. Specifically, the VL tasks cover image-text retrieval (\textbf{ITR}: MSCOCO~\cite{Datasets:MSCOCO}, Flickr30K~\cite{Datasets:Flickr30k}), video-text retrieval (\textbf{VTR}: MSR-VTT~\cite{Datasets:MSRVTT}, MSVD~\cite{Datasets:MSVD}), question answering (\textbf{VQA}\&\textbf{GQA}: VQAv2~\cite{Datasets:VQAv2}, GQA~\cite{Datasets:GQA}), and visual grounding (\textbf{VG}: RefCOCO, RefCOCO+~\cite{Datasets:REFCOCO}, RefCOCOg~\cite{Datasets:REFCOCOG}). 
For ITR and VTR, we report Recall@1 (R@1) on sentence retrieval (I-T, V-T), image retrieval (T-I), and video retrieval (T-V), and Rsum of R@1,5,10 in two directions for comprehensive verification. For VQA and GQA, we compare the results on Test-Dev and Test-Std sets, while for VG, we report the performance on the images containing multiple people (TestA) and multiple instances of all other objects (TestB). The GLUE benchmark~\cite{Datasets:GLUE} consists of linguistic acceptability (CoLA~\cite{Datasets:CoLA}), sentiment analysis (SST2~\cite{Datasets:SST-2}), similarity and paraphrase (MRPC~\cite{Datasets:MRPC}, QQP~\cite{Datasets:QQP}, STS-B~\cite{Datasets:STS-B}), and natural language inference (MNLI~\cite{Datasets:MNLI}, QNLI~\cite{Datasets:QNLI}, RTE~\cite{Datasets:RTE}). More details are shown in the Appendix.

\textbf{Training Details.} 
All experiments are conducted using eight GeForce RTX 3090Ti (24GB) and keep most default configurations of the pre-trained models, \eg, choice of optimizer, warm-up schedule, input image resolution, video sequence length, input text processing, \etc. Specially, we follow their original batch sizes for most tasks, with the exception of maximum batch size$=$112 (vs. raw 128) for VSE$\infty$ in ITR, due to the out-of-memory (OOM) issue. The reduction factors $r$ of LST/UniPT are set to 8 and 4/2 for NLP and VL tasks respectively, unless otherwise noted. We also maintain the original learning rate $lr$ and the number of epochs for fully fine-tuning, and search learning rates over $\left\{20\times lr,10\times lr,lr\right\}$ for PETL methods.
More training details and hyper-parameters are shown in the Appendix.

\addtolength{\tabcolsep}{-3pt}
\begin{table*}[t]
\begin{minipage}[t]{\textwidth}
    \centering
    \resizebox{\textwidth}{!}{
    \begin{tabular}{lcccccccccccc|ccccccccc}
    \toprule
    \multirowcell{2}{Method} 
    &Params.
    &\multicolumn{2}{c}{Memory (G)}
    &\multicolumn{3}{c}{Flickr30K} 
    &\multicolumn{3}{c}{MSCOCO1K} 
    &\multicolumn{3}{c|}{MSCOCO5K}
    &Params.
    &\multicolumn{2}{c}{Memory (G)}
    &\multicolumn{3}{c}{MSR-VTT} 
    &\multicolumn{3}{c}{MSVD}
    \\  
    \cmidrule{5-13} \cmidrule{17-22}
    &(M) &Train &Test 
    &I-T &T-I &Rsum 
    &I-T &T-I &Rsum
    &I-T &T-I &Rsum
    &(M) &Train &Test
    &T-V &V-T &Rsum 
    &T-V &V-T &Rsum
    \\
    \midrule
    \gray{Fully-FT} 
    &\gray{201.2} &\gray{22.1 * 8} &\gray{20.12}
    &\gray{85.6} &\gray{73.3} &\gray{546.6} 
    &\gray{83.1} &\gray{71.7} &\gray{542.7}
    &\gray{64.2} &\gray{51.2} &\gray{468.9} 
    &\gray{151.3} &\gray{12.2 * 4} &\gray{1.12}
    &\gray{42.8} &\gray{42.1} &\gray{389.2} 
    &\gray{45.2} &\gray{57.1} &\gray{425.5}
    \\
    \gray{Partially$\Downarrow$} 
    &\gray{2.9} &\gray{15.0} &\gray{20.12}
    &\gray{75.6} &\gray{59.8} &\gray{508.3}
    &\gray{74.6} &\gray{59.5} &\gray{510.5}
    &\gray{51.2} &\gray{37.6} &\gray{401.7} 
    &\gray{0.7} &\gray{1.9 * 4} &\gray{1.12}
    &\gray{36.4} &\gray{37.0} &\gray{353.9}
    &\gray{37.4} &\gray{52.4} &\gray{406.4}
    \\
    LST~\cite{TL:LST}
    &9.7 &15.1 &20.31
    &82.1 &66.5 &529.5
    &78.2 &64.8 &525.8
    &57.8 &43.1 &434.5 
    &11.2 &8.0 * 4 &1.15
    &37.0 &37.8 &356.7
    &35.5 &55.4 &407.2
    \\
    \textbf{UniPT} 
    &12.4 &15.1 &20.19
    &\textbf{84.8} &\textbf{69.1} &\textbf{537.4}
    &\textbf{80.6} &\textbf{67.5} &\textbf{532.9}
    &\textbf{61.1} &\textbf{45.9} &\textbf{445.3}
    &9.6 &3.4 * 4 &1.13
    &\textbf{38.9} &\textbf{39.3} &\textbf{361.3}
    &\textbf{40.9} &\textbf{59.7} &\textbf{432.1}
    \\
    \end{tabular}
    }
\end{minipage} \\
\begin{minipage}[t]{\textwidth}
    \centering
    \resizebox{\textwidth}{!}{
    \begin{tabular}{lccccccc|ccccccccccc}
    \toprule
    \multirowcell{2}{Method} 
    &Params. 
    &\multicolumn{2}{c}{Memory (G)}
    &\multicolumn{2}{c}{VQAv2} 
    &\multicolumn{2}{c|}{GQA}
    &Params.
    &\multicolumn{2}{c}{Memory (G)}
    &\multicolumn{3}{c}{RefCOCO} 
    &\multicolumn{3}{c}{RefCOCO+} 
    &\multicolumn{2}{c}{RefCOCOg}
    \\  
    \cmidrule{5-8} \cmidrule{12-19}
    &(M) &Train &Test
    &Test$_\text{Dev}$ &Test$_\text{Std}$
    &Test$_\text{Dev}$ &Test$_\text{Std}$
    &(M) &Train &Test
    &Val &TestA &TestB 
    &Val &TestA &TestB
    &Val &Test 
    \\
    \midrule
    \gray{Fully-FT} 
    &\gray{236.8} &\gray{20.5 * 4} &\gray{12.64}
    &\gray{76.71} &\gray{76.86} &\gray{60.25} &\gray{61.44} 
    &\gray{185.2} &\gray{19.8 * 2} &\gray{3.36}
    &\gray{86.51} &\gray{89.13} &\gray{81.22}
    &\gray{79.54} &\gray{84.54} &\gray{70.63}
    &\gray{80.92} &\gray{80.95}
    \\
    \gray{Partially$\Uparrow$}  
    &\gray{117.6} &\gray{10.5 * 4} &\gray{12.64}
    &\gray{76.73} &\gray{76.84} &\gray{61.13} &\gray{62.28}
    &\gray{18.3} &\gray{11.3 * 2} &\gray{3.36} 
    &\gray{85.39} &\gray{88.40} &\gray{79.78}
    &\gray{77.66} &\gray{84.00} &\gray{69.38}
    &\gray{79.92} &\gray{80.08}
    \\
    LST~\cite{TL:LST}
    &13.4 &6.4 * 4 &12.76
    &75.29 &75.44 &59.93 &\textbf{60.75}
    &0.9 &6.3 * 2 &3.41
    &81.63 &85.19 &76.03
    &71.32 &78.20 &62.06
    &72.53 &73.67
    \\
    \textbf{UniPT} 
    &10.3 &2.9 * 4 &12.67
    &\textbf{75.33} &\textbf{75.53} &\textbf{60.10} &60.72
    &0.7 &3.4 * 2 &3.38
    &\textbf{82.71} &\textbf{86.25} &\textbf{78.16}
    &\textbf{72.94} &\textbf{79.18} &\textbf{64.49}
    &\textbf{77.04} &\textbf{77.33}
    \\
    \bottomrule
    \end{tabular}
    }
\end{minipage}
\vspace{-0.8em}
\caption{Comparisons with the best memory-efficient counterpart LST on VL tasks with various architectures. We adopt \emph{VSE$\infty$} with both CNN and Transformer {\scriptsize(ResNeXt-101 + BERT-base)} on ITR task, \emph{CLIP4Clip} with dual Transformer encoders {\scriptsize(ViT-base + Text Transformer)} on VTR task, \emph{CLIP-ViL} with Cross-modal Transformer on VQA and GQA tasks, and \emph{MDETR} with Encoder-Decoder architecture on VG task.}
\label{tab:VL_METL}
\end{table*}
\addtolength{\tabcolsep}{3pt}

\subsection{Main Results}
\label{subsec:mainresult}

\subsubsection{UniPT vs. Non-memory-efficient PETL}
\textbf{Model Settings.} 
In~\Cref{tab:GLUE_PETL,tab:VL_PETL}, we compare our UniPT against Fully Fine-Tuning and several popular PETL methods. In \textit{Adapter}~\cite{TL:Adapter-BERT}, we insert tiny trainable modules into every attention and feed-forward layer of the Transformer and conv2-5 layers of CNN, while in \textit{BitFit}~\cite{TL:BitFit}, we only update the bias terms for the pre-trained models. Besides, \textit{LoRA}~\cite{TL:LoRA} and \textit{Prompt}~\cite{TL:Prefix-Tuning} introduce additional trainable low-rank matrices and learnable token inputs into the attention layer of the Transformer that are not compatible with CNN. We reproduce \textit{LST}~\cite{TL:LST} by extracting the ladder blocks from the pre-trained networks, but for CNN, we abandon the corresponding conv1 layer in the side network to address the size misalignment across layers. In Fully Fine-Tuning (\textit{Fully-FT}), all network parameters are updated, while in Partially Tuning (\textit{Partially$\Downarrow$}), we only update the last projection and aggregation modules after the pre-trained network.

\noindent\textbf{Results on VL Tasks.}
In~\Cref{tab:VL_PETL}, we select various transfer paradigms for diverse and challenging validation. To identify the impact on pre-trained CNN, single Transformer, and dual Transformer encoders, we adopt two structures from \emph{VSE$\infty$} on ITR and one construction from \emph{CLIP4Clip} on VTR:
\begin{itemize}
    \item ResNeXt-101(32×8d)~\cite{CNN:ResNeXt17} + BiGRU~\cite{RNN:Bi-GRU} (tiny);
    \item BERT-base~\cite{TransF:BERT} + BUTD regions~\cite{IC:BU_TDA} (tiny);
    \item ViT-base~\cite{TransF:ViT} + Text Transformer~\cite{VLP:CLIP}.
\end{itemize}
We can discover that our UniPT achieves competitive results with recent state-of-the-art PETL works including SSF~\cite{TL:SSF}, FacT~\cite{TL:FacT}, and AdaLoRA~\cite{TL:AdaLoRA}, and even outperforms some popular PETL methods in a low-memory regime. 
Note that \textit{(1) Adapter displays a difficult optimization situation on CNN and achieves sub-optimal performance after careful adjustments.} We assume that this may be due to the CNN's deeper network structure compared with the Transformer, leading to the vanishing gradient problem of the Adapter inside the shallow layers.
\textit{(2) For CNN, the two-stage interaction of UniPT leads to a modest increase in trainable parameters, and the marginal memory gains are primarily a result of reaching memory-saving saturation.} That is because the forward process of the large backbone takes up the main memory costs, where the simplest transfer method Partially$\Downarrow$ still occupies the memory overhead by 14.9G.

\noindent\textbf{Results on NLP Benchmark.} 
\Cref{tab:GLUE_PETL} demonstrates the comparisons on GLUE benchmark. Based on the \emph{T5-base}, UniPT significantly reduces memory overhead from 17.6G to 2.9G with similar trainable parameter usage, and achieves competitive performance as full fine-tuning and other PETL works. To further utilize the memory-efficient advantage of UniPT, we cooperate it with \emph{T5-large} and find that even with lower memory budget in Adapter, LoRA, and BitFit, UniPT with can surpass other PETL methods by a large margin. \textit{Limited by current device, we would accommodate the training of UniPT on the T5-3B model in the future.}

\subsubsection{UniPT vs. Memory-efficient PETL}
\textbf{Model Settings.} 
In~\Cref{tab:VL_METL}, we compare our UniPT against the best memory-efficient competitor LST. 
For \emph{VSE$\infty$}~\cite{ITM:GPO} on ITR task and \emph{CLIP4Clip}~\cite{VLP:CLIP4Clip} on VTR task, we freeze their dual encoders and only update the last small aggregation modules during training as \textit{Partially$\Downarrow$}, which serves as the lowest performance bound for LST and UniPT. 
Besides for \emph{CLIP-ViL}~\cite{VLP:CLIP-ViL} on QA task and \emph{MDETR}~\cite{VLP:MDETR} on VG task, we freeze the vision backbone except for \textit{Fully-FT}, because the performance and efficiency gains of fine-tuning it over keeping it frozen are both limited~\cite{TL:VL-ADAPTER,TL:LST}. Meanwhile, we unfreeze their respective cross-modal or encoder-decoder Transformer during fine-tuning as \textit{Partially$\Uparrow$}, which thereby represents the upper performance bound for LST and UniPT.

\noindent\textbf{Results on VL Tasks.}
\Cref{tab:VL_METL} shows the comparisons on diverse VL tasks with various pre-trained architectures:
\begin{itemize}
    \item ITR task: \emph{VSE$\infty$}~\cite{ITM:GPO} with the strongest combination of BERT-base~\cite{TransF:BERT} model and ResNeXt-101(32×8d)~\cite{CNN:ResNeXt17} backbone pre-trained on Instagram (WSL)~\cite{Datasets:WSL};
    \item VTR task: \emph{CLIP4Clip}~\cite{VLP:CLIP4Clip} with pre-trained CLIP~\cite{VLP:CLIP} using Text Transformer~\cite{TransF:GPT-2} and ViT-B/32~\cite{TransF:ViT} models;
    \item QA task: \emph{CLIP-ViL}~\cite{VLP:CLIP-ViL} that utilizes CLIP image backbone~\cite{VLP:CLIP} and encodes the text into word embedding sequence, followed by a cross-modal Transformer;
    \item VG task: \emph{MDETR}~\cite{VLP:MDETR} with pre-trained ResNet-101~\cite{CNN:Resnet16}, RoBERTa-base~\cite{TransF:RoBERTa}, and encoder-decoder Transformer.
\end{itemize}
\noindent \textit{(1) Larger performance gains.} 
Our UniPT outweighs the best LST on various tasks and backbones. In particular, for the larger and more compelling retrieval datasets \ie MSCOCO5K and MSR-VTT, our UniPT achieves absolute R@1 improvements than LST by 3.3/2.8\% and 1.9/1.5\%, validating the superior applicability in handling more challenging matching patterns.
Besides, on the much smaller MSVD dataset, our UniPT obtains much better Rsum than Fully-FT (432.1 vs. 425.5\%), indicating the anti-overfitting capability under the limited data-driven scenario.
\textit{(2) Lower training memory usage.} UniPT achieves nearly twice the training memory savings of LST in most cases, excluding the \emph{VSE$\infty$} in ITR and \emph{T5-large} in GLUE. 
\textit{(3) Negligible inference memory costs.} UniPT requires less extra memory consumption during the inference process by an average of 0.02G vs. 0.06G of LST with various networks on all multi-modal tasks.
In summary, our UniPT outperforms the leading counterpart LST by achieving the optimal trade-off in both training efficiency and transfer capability.

\begin{figure*}[t]
  \begin{minipage}[t]{0.40\textwidth}
    \centering
    \includegraphics[width=\linewidth,trim= 15 270 340 0,clip]{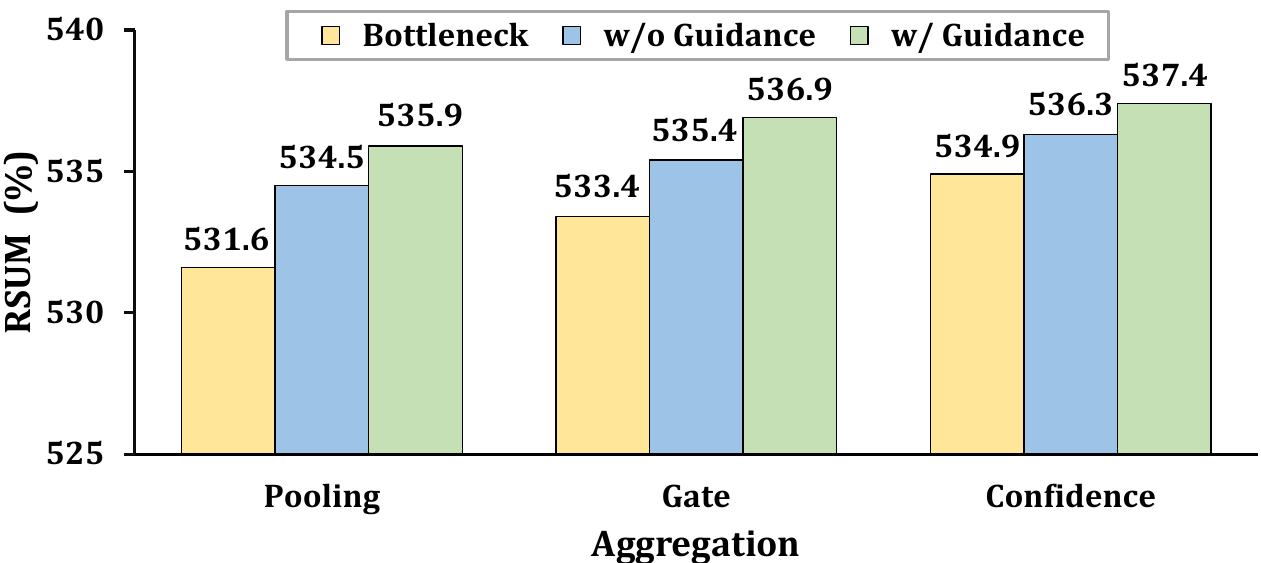}
    \vspace{-1.8em}
    \caption{Rsum (\%) on Flickr30K by \emph{VSE$\infty$} w/ or w/o interaction guidance and dynamic aggregation strategies.}
    \label{fig:guidance_attention}
  \end{minipage}
  \hfill%
  \begin{minipage}[t]{0.32\textwidth}
    \centering
    \includegraphics[width=\linewidth,trim= 15 270 440 0,clip]{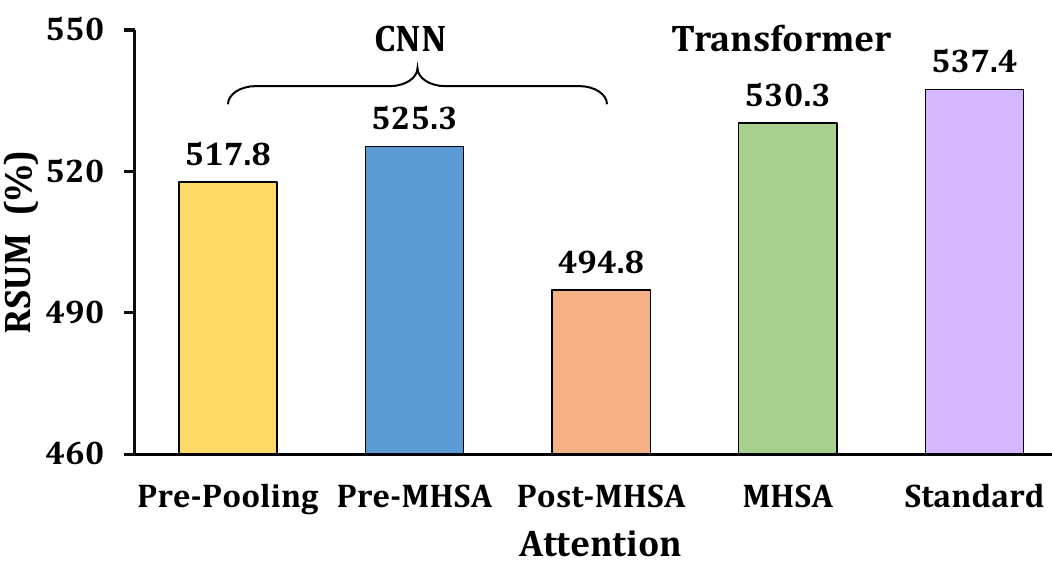}
    \vspace{-1.8em}
    \caption{Rsum (\%) on Flickr30K by \emph{VSE$\infty$} w/ our truncated or other alternative attention.}
    \label{fig:hierarchical_attention}
  \end{minipage}
  \hfill%
  \begin{minipage}[t]{0.25\textwidth}
    \centering
    \includegraphics[width=\linewidth,trim= 15 270 550 0,clip]{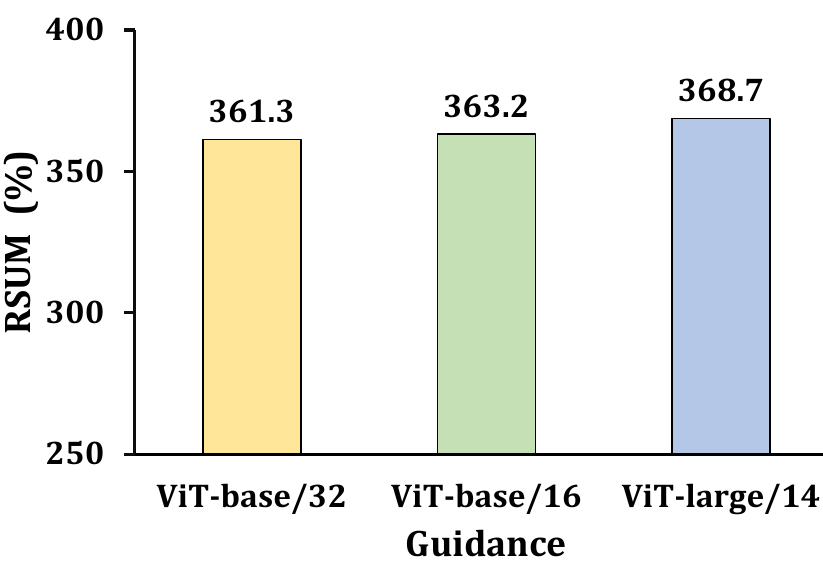}
    \vspace{-1.8em}
    \caption{Rsum (\%) on MSR-VTT by {\footnotesize\emph{CLIP4Clip}} w/ stronger guidance.}
    \label{fig:guidance_feature}
  \end{minipage}
  \vspace{-1em}
\end{figure*}

\begin{figure}[t]
    \centering
    \includegraphics[width=0.90\linewidth,trim= 15 260 400 0,clip]{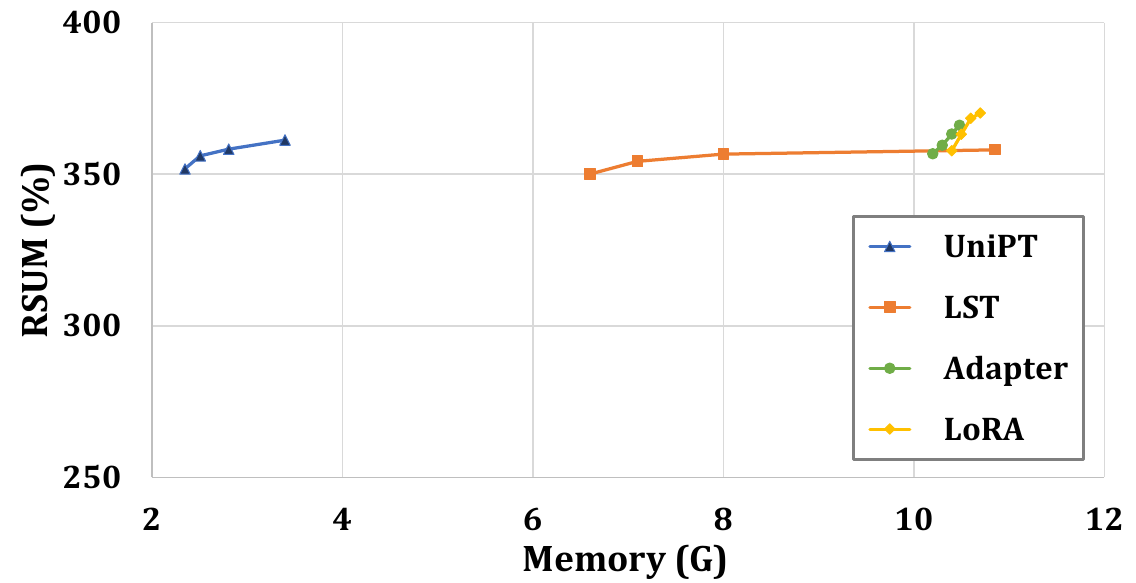}
    \vspace{-0.5em}
    \caption{Accuracy-memory trade-off on MSR-VTT by {\footnotesize\emph{CLIP4Clip}} for several PETL approaches w/ reduction factor $r\in \{8,4,2,1\}$.}
    \label{fig:reduction_factor}
\end{figure}

\subsection{Ablation Studies}
\label{subsec:ablation}

\textbf{Necessity of Interaction Guidance and Dynamic Fusion.}

\noindent\textbf{\emph{Settings.}} \Cref{fig:guidance_attention} compares several UniPT variations using \emph{VSE$\infty$} on Flickr30K. \textbf{(1)} For interaction, we test Adapter-like network (Bottleneck) and self-attention between intermediates inside each layer (w/o Guidance) as counterparts. \textbf{(2)} For aggregation, we utilize average pooling (Pooling) and static ladder-gated connection (Gate) for comparison.

\noindent\textbf{\emph{Results.}} In~\Cref{fig:guidance_attention}, we discover that interaction w/ Guidance obtains more promising performance than the ones w/o Guidance, indicating that guidance from the last layer output could enhance hidden features inside each layer with more powerful adaptation capability. 
Besides, dynamic confidence also suppresses other strategies by adaptively adjusting the proportion of each layer over various inputs and producing more discriminative representations for domain transfer.

\textbf{Superiority of UniPT over Multi-head Self-attention.} 

\noindent\textbf{\emph{Settings.}} \Cref{fig:hierarchical_attention} reports the comparisons between our truncated attention (Standard) and multi-head self-attention (MHSA) using \emph{VSE$\infty$} on Flickr30K. We set the head of MHSA as 4 (work best). Here, we replace our standard attention in the pre-interaction layer in CNN by Pooling (Pre-Pooling CNN) and MHSA (Pre-MHSA CNN), and the post-interaction layer by MHSA in CNN (Post-MHSA CNN). 

\noindent\textbf{\emph{Results.}} Current UniPT attention outperforms MHSA by a large margin in all alternative locations. Notably, Post-MHSA in CNN shows a sharp decline in accuracy that reflects its incompatibility and instability over various complex interaction patterns for feature maps or token embeddings.

\textbf{Bonus of Stronger Backbone Output as Guidance.} 

\noindent\textbf{\emph{Settings.}} In~\Cref{fig:guidance_feature}, we take larger ViT-B/16 or ViT-L/14 as stronger guidance for our UniPT with ViT-B/32 using \emph{CLIP4Clip} on MSR-VTT. The larger model does not directly participate in the final output, which only serves as queries to compute interaction and confidence weights for ViT-B/32. 

\noindent\textbf{\emph{Results.}} We surprisingly find that more powerful guidance brings more beneficial gains, further validating the significance of introducing interaction and aggregation guidance.

\textbf{Optimal Balance of UniPT over PETL Methods.}

\noindent\textbf{\emph{Settings.}} \Cref{fig:reduction_factor} shows accuracy-memory trade-off using \emph{CLIP4Clip} on MSR-VTT with varying reduction factor $r$. 

\noindent\textbf{\emph{Results.}} UniPT stands out by dramatically reducing training memory usage with competitive performance as LoRA, Adapter, and LST. Besides, it displays good stability and robustness across a diverse spectrum of side network sizes.


\section{Conclusion}
In this paper, we propose Universal Parallel Tuning (UniPT), a brand-new PETL paradigm that hits a sweet spot between performance and memory efficiency on various VL and GLUE downstream tasks. 
Crucially, our research unveils two intriguing phenomena: 1) Delicately leveraging the ultimate outputs as guidance can facilitate the adaptation capability of feature representations inside each layer of the pre-trained network; 2) Dynamically learning the aggregation weights also demonstrates optimal strategy and greater suitability for various inputs across modalities.
By incorporating a lightweight network in parallel, our UniPT capitalizes on various pre-trained architectures across diverse domains, and more importantly, requires no backpropagation through the large backbone network. Extensive experiments have validated that our UniPT can not only significantly reduce memory footprint and surpass existing memory-efficient methods with good flexibility and broad applicability over various network backbones, but also achieve impressive benefits beyond recent PETL methods in a low-memory regime.

\noindent\textbf{Limitations.} We believe UniPT is a brand-new attempt for a desired and powerful architecture capable of memory-efficient transfer learning over various pre-trained backbones. A key limitation is that there is still a performance gap between UniPT and fully fine-tuning. Besides, a large input size (\eg, shallow feature maps in CNN) may affect the computational complexity of UniPT, and the current pre-post design can further alleviate the memory consumption but improve the trainable parameters to some extent. Limited by current resources, we are committed to developing UniPT for larger fundamental models (\eg, LLMs) in the future.

\begin{appendix}


\begin{table*}[t]
    \centering
    \resizebox{\textwidth}{!}{
    \begin{tabular}{lccc}
    \toprule
    \makecell{Method} 
    &\makecell{Learning rate} 
    &\makecell{Batch size}
    &\makecell{Other hyperparameters} \\             
    \midrule
    Full fine-tuning 
    & $3 \times 10^{-4}$ & 100 & -- \\
    Adapters  
    & $3 \times 10^{-4}$ & 100 & hidden dimension=48\\
    LoRA  
    & $3 \times 10^{-4}$ & 100 & rank=32 \\
    BitFit  
    & $3 \times 10^{-4}$ & 100 & -- \\
    Prompt-tuning 
    & $3 \times 10^{-1}$ & 100 & prompt number=100 \\
    \midrule
    LST / UniPT 
    & $3 \times 10^{-3}$ & 100 & \makecell{r=8; kept layer index=1,2,3,5,6,7,9,10,11}\\
    LST / UniPT (T5-large) 
    & $3 \times 10^{-3}$ & 100 & \makecell{r=8; kept layer index=1,3,5,7,9,11,13,15,17,19,21,23} \\
    \bottomrule
    \end{tabular}}
    \caption{Hyperparameters used for the GLUE benchmark.}
    \label{tab:hyperparameters-glue-tasks}
\end{table*}

\begin{table*}[t]
    \centering
    \resizebox{\textwidth}{!}{
    \begin{tabular}{lrrrrrrrr}
    \toprule
    \multirowcell{2}{Hyperparameters}
    &\multicolumn{2}{c}{MSCOCO / Flickr30K}
    &\multicolumn{2}{c}{MSR-VTT / MSVD}
    &\multicolumn{2}{c}{VQA / GQA}
    &\multicolumn{2}{c}{RefCOCO / + / g} \\
    &Fully-FT &PETL
    &Fully-FT &PETL
    &Fully-FT &PETL
    &Fully-FT &PETL \\
    \midrule
    Peak learning rate 
    &5e-4 &5e-4 &1e-4 &1e-4 &5e-5 &5e-4 &5e-5 &5e-4\\
    Total epochs 
    &25 &25 &5 &5 &5 &5 &10 &10\\
    Warmup 
    &linear &linear &cosine &cosine
    &linear &linear &linear &linear\\
    Batch size 
    &112 &112 &128 &128 &256 &256 &8 &8\\
    AdamW $\beta_{1}$ 
    &0.9 &0.9 &0.9 &0.9 &0.9 &0.9 &0.9 &0.9\\
    AdamW $\beta_{2}$
    &0.999 &0.999 &0.98 &0.98
    &0.999 &0.999 &0.999 &0.999\\
    Weight decay 
    &1e-4 &1e-4 &0.2 &0.2 
    &0.01 &0.01 &1e-4 &1e-4\\
    Vision resolution 
    &512$^{2}$ &512$^{2}$ &12*224$^{2}$ &12*224$^{2}$
    &384*640 &384*640 &raw size &raw size\\
    \bottomrule
    \end{tabular}}
    \caption{Hyperparameters used for ITR, VTR, VQA, GQA, and VG tasks.}
    \label{tab:hyperparameters-vl-tasks}
\end{table*}

\section{Appendix}

\noindent\textbf{Ethics Statement.} We propose a Universal Parallel Tuning for parameter- and memory-efficient fine-tuning of popular pre-trained networks in NLP and VL tasks. These fundamental frameworks offer numerous positive applications, \eg sentiment analysis, natural language inference, cross-modal retrieval, visual question answering, visual grounding, and \etc. However, it is important to acknowledge the ethical concerns associated with existing large pre-trained models. One such concern involves the potentially biased or discriminative information in the mass training data~\cite{ES:ETD_LLM}. Additionally, the leakage of private information from training data has been identified as another significant issue~\cite{ES:IRGB}. Furthermore, the environmental impact stemming from the training processes has also been a subject of discussion~\cite{ES:EPC_NLP}.

Our UniPT aims to train large pre-trained models with minimal alterations to their pre-existing parameters. To our best knowledge, no previous work has examined the topic of whether the parameter-efficient transfer learning (PETL) methods amplify, alleviate, or have negligible effects on these concerns, \eg bias or information leakage, which deserve further exploration. Compared with fully fine-tuning (mainstream practice), our methods drastically reduce the number of trainable parameters, especially the memory occupation of GPU devices, and have a negligible negative impact on training and inference overheads, which would significantly reduce the deployment cost of serving models in terms of memory and servers. Last but not least, most of the training experiments in this paper were conducted in a data center powered entirely by the renewable energy.

\begin{table*}[t]
    \centering
    \resizebox{\textwidth}{!}{
    \begin{tabular}{lccccccccc}
    \toprule
    Method &MSCOCO &Flickr30K &MSR-VTT &MSVD &VQA &GQA &RefCOCO &RefCOCO+ &RefCOCOg \\
    \midrule
    Fully-FT 
    &2.5d (8gpu) &17.5h (8gpu) &30.6h &11.5h &9.6h &15.6h &34.2h &36.6h &28.3h\\
    Partially 
    &3.0d (1gpu) &18.5h (1gpu) &5.7h &2.1h &3.9h &5.8h &19.5h &20.4h &12.6h \\
    LST 
    &4.1d (1gpu) &25.8h (1gpu) &11.2h &5.2h &3.5h &4.8h &19.1h &20.0h &12.2h \\
    UniPT 
    &3.4d (1gpu) &20.9h (1gpu) &7.1h &3.5h &2.9h &3.7h &18.6h &19.7h &11.5h \\
    \midrule
    Fully-FT &28.18m&5.5m&5.6m&2.8h&3.3h&23.5h&8.4h&10.8h&3.2h\\
    Partially &28.18m&5.5m&5.6m&2.8h&3.3h&23.5h&8.4h&10.8h&3.2h\\
    LST &61.42m&7.8m&10.1m&3.6h&4.1h&29.4h&10.5h&13.5h&4.3h\\
    UniPT
    &42.87m &6.1m &9.2m &3.1h&3.7h &26.7h &9.0h &12.1h &3.7h\\
    \bottomrule
    \end{tabular}
    }
    \caption{Training and inference time on various VL datasets. The top and bottom indicate the training and inference process respectively.}
    \label{tab:training-inference-time}
\end{table*}

\subsection{Setups}

\textbf{Training Details.}
In \Cref{tab:hyperparameters-glue-tasks}, we provide an overview of the training hyperparameters for both PETL and our methods when used in conjunction with the \emph{T5-series}~\cite{TransF:T5} framework on the GLUE benchmark.
In~\Cref{tab:hyperparameters-vl-tasks}, we outline the training hyperparameters for \emph{VSE$\infty$}~\cite{ITM:GPO}, \emph{CLIP4Clip}~\cite{VLP:CLIP4Clip}, \emph{CLIP-ViL}~\cite{VLP:CLIP-ViL}, and \emph{MDETR}~\cite{VLP:MDETR} on various VL tasks.

\textbf{Model Architectures.}
\textit{1) For the GLUE benchmark,} we follow the publicly available code of LST~\cite{TL:LST} to keep the same configurations between UniPT and other PETL methods.
\textit{2) For the VL tasks,} we reproduce the \emph{VSE$\infty$}~\cite{ITM:GPO}, \emph{CLIP4Clip}~\cite{VLP:CLIP4Clip}, \emph{CLIP-ViL}~\cite{VLP:CLIP-ViL}, and \emph{MDETR}~\cite{VLP:MDETR} and maintain their original network configurations according to their publicly released codes. Notably, we implement all the parameter-efficient tuning methods (\ie Adapter, BitFit, LoRA, Prompt) by the commonly-used AdapterHub library~\cite{TL:AdapterHub}, and memory-efficient LST from its official GitHub source~\cite{TL:LST}. Following their optimal recommendations, we set the reduction factor and nonlinear activation as 16 and ReLU respectively for \textbf{Adapter} after each self-attention and feed-forward layer. For \textbf{LoRA}, we set the rank and alpha as 8 and 16, while for \textbf{Prompt}, we set the number of extra prompts as 30. Besides, we train all the bias items for \textbf{BitFit}, while for \textbf{LST}, we set the reduction factor $r$ as 4 and only drop part of the side layers for large encoder-decoder backbone following its standard settings.

\textbf{Datasets.} 
For the GLUE benchmark~\cite{Datasets:GLUE}, we implement UniPT on eight NLP tasks including linguistic acceptability (CoLA~\cite{Datasets:CoLA}), sentiment analysis (SST2~\cite{Datasets:SST-2}), similarity and paraphrase (STS-B~\cite{Datasets:STS-B}, MRPC~\cite{Datasets:MRPC}, QQP~\cite{Datasets:QQP}), and natural language inference (RTE~\cite{Datasets:RTE}, QNLI~\cite{Datasets:QNLI}, MNLI~\cite{Datasets:MNLI}). 
We report the accuracy metric for SST-2, MNLI, QNLI, and RTE. 
For CoLA and STS-B, we present Matthew's Correlation and Pearson-Spearman Correlation as evaluation metrics, respectively. 
For MRPC and QQP, we show average results of F1 score and accuracy. Besides, we report the maximum memory usage during training and evaluation on RTE. Each result is an average obtained from three different seeds.
 
For Image-Text Retrieval (ITR), we conduct experiments on MSCOCO~\cite{Datasets:MSCOCO} and Flickr30K~\cite{Datasets:Flickr30k}. Each image is corresponding to five text captions. 
MSCOCO contains 123,287 images, while Flickr30K contains 31,000 images. Following the data split of~\cite{ITM:VSE++}, we divide the MSCOCO dataset into 113,287 training images, 5000 validation images, and 5000 test images, while there are 29,000 training images, 1000 validation images, and 1000 test images on Flickr30K dataset. 
Note that the evaluation results on MSCOCO are reported in 5K and 1K test sets respectively, where the 1K reported results are averaged over the five 1K data folds.

For Video-Text Retrieval (VTR), we validate our method on MSR-VTT~\cite{Datasets:MSRVTT} and MSVD~\cite{Datasets:MSVD}. 
MSR-VTT consists of 10,000 videos, each ranging
from 10 to 32 seconds, and 200,000 captions. Following the data
splits from~\cite{VLP:MMT}, we adopt 1k-A protocol that utilizes
9,000 videos with all corresponding captions for training and utilizes 1,000 video-text pairs for testing.
Besides, MSVD includes 1,970 videos with about 80,000 captions, where train, validation, and test splits are divided into 1200, 100, and 670 videos. Here, we report the test results with multiple captions per video.

For Visual and Compositional Question Answering (VQA \& GQA), we test our method on VQAv2~\cite{Datasets:VQAv2} and GQA~\cite{Datasets:GQA}. VQAv2 is a dataset constructed from MSCOCO~\cite{Datasets:MSCOCO}, which has 83k/41k/81k images and 443k/214k/453k question-image pairs for training/validation/testing.
Besides, GQA consists of 113K images and 22M questions generated from ground truth image scene graph, measuring performance on an array of reasoning skills, \eg object and attribute recognition, spatial reasoning, \etc. Note that we report the performance on test-dev and test-standard splits.

For Visual Grounding (VG), we adopt RefCOCO, RefCOCO+~\cite{Datasets:REFCOCO}, and RefCOCOg~\cite{Datasets:REFCOCOG} for testing. 
RefCOCO has 142,210 referring expressions for 50,000 bounding boxes in 19,994 images from MSCOCO. These are divided into the train, validation, Test A, and Test B with 120,624, 10,834, 5,657, and 5,095 samples, respectively.  The categories of bounding boxes in TestA are people while the ones in TestB are objects.
RefCOCO+ has 141,564 expressions for 49,856 boxes in 19,992 images from MSCOCO that are divided into the train (120,191), val (10,758), Test A (5,726) and Test B (4,889) splits. Compared to RefCOCO, its expressions include more appearances (attributes) than absolute locations.
RefCOCOg has 104,560 expressions for 54,822 objects in 26,711 images. Compared to RefCOCO and RefCOCO+, its expressions are collected in a non-interactive way, and the lengths of contents are longer (8.4 words on average).

\begin{figure*}[t]
    \begin{minipage}[t]{0.48\textwidth}
        \centering
        \includegraphics[width=\linewidth, height=0.65\linewidth,trim= 0 200 450 0,clip]{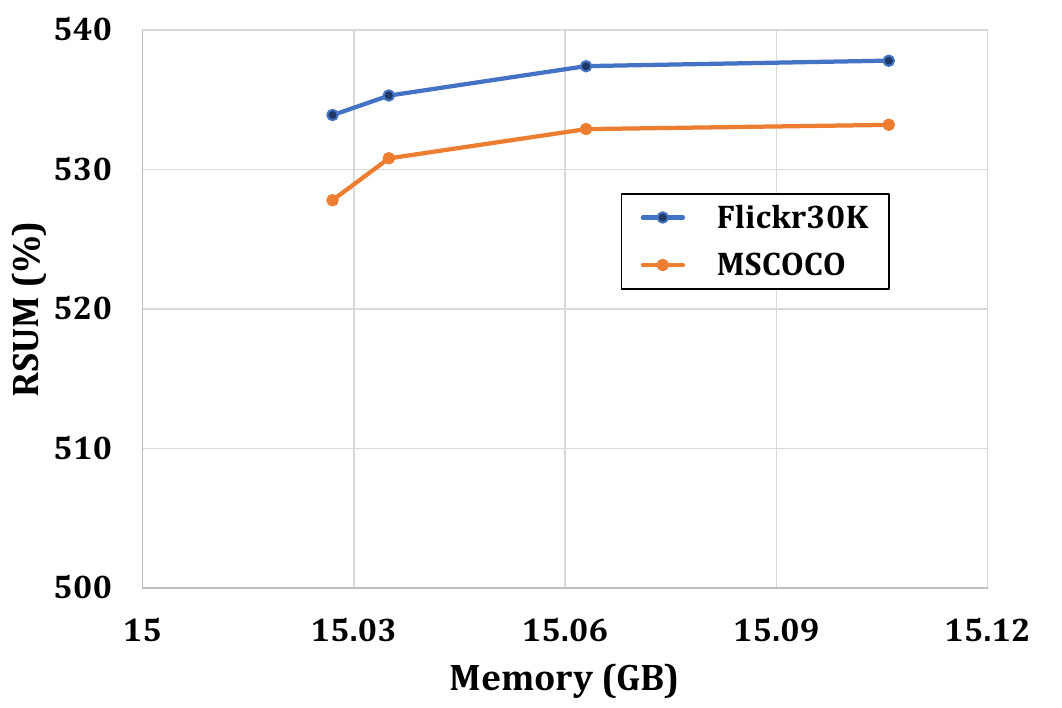}
        \caption{Impact of reduction factor $r\in \{8,4,2,1\}$ on ITR.}
        \label{fig:reduction_factor_ITR}
    \end{minipage}
    \hfill%
    \begin{minipage}[t]{0.48\textwidth}
        \centering
        \includegraphics[width=\linewidth, height=0.65\linewidth,trim= 0 200 450 0,clip]{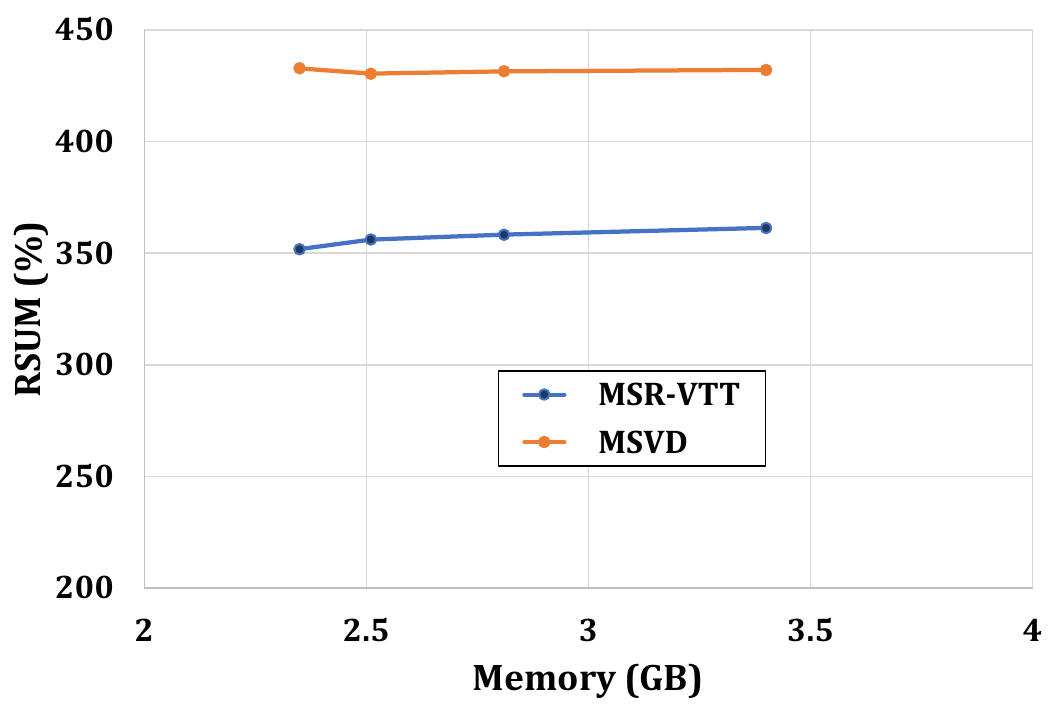}
        \caption{Impact of reduction factor $r\in \{8,4,2,1\}$ on VTR.}
        \label{fig:reduction_factor_VTR}
        \end{minipage} \\
    \begin{minipage}[t]{0.48\textwidth}
        \centering
        \includegraphics[width=\linewidth, height=0.65\linewidth,trim= 0 200 400 0,clip]{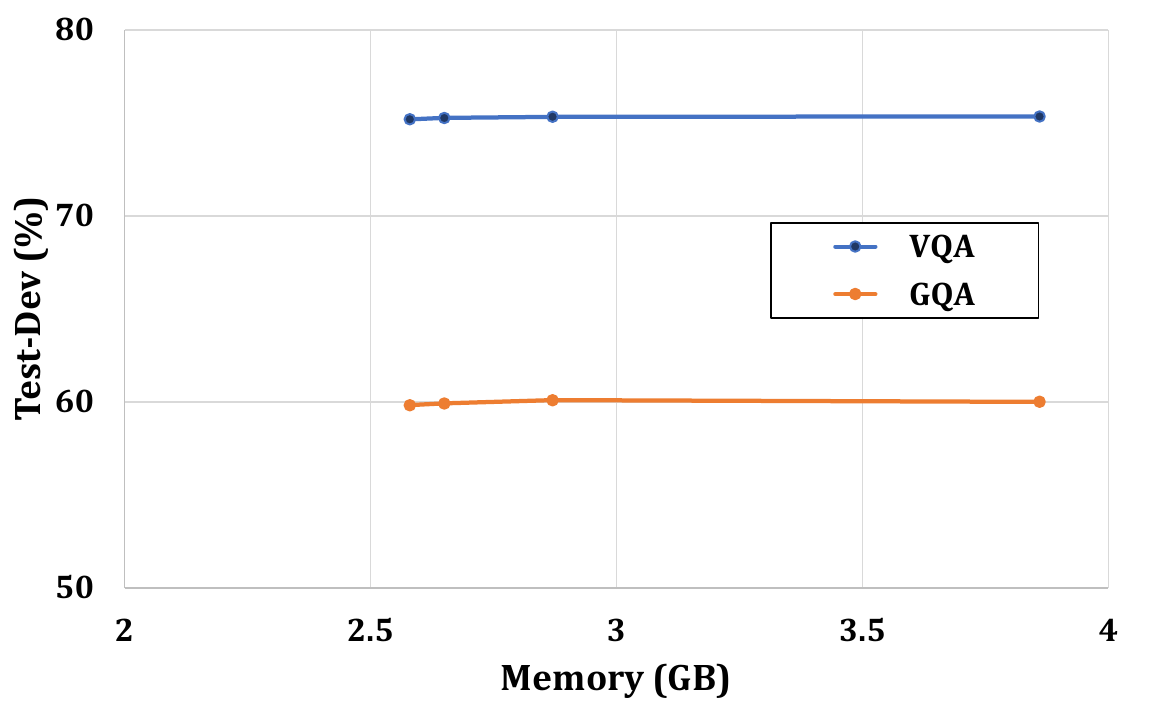}
        \caption{Impact of reduction factor $r\in \{8,4,2,1\}$ on QA.}
        \label{fig:reduction_factor_QA}
    \end{minipage}
    \hfill%
    \begin{minipage}[t]{0.48\textwidth}
        \centering
        \includegraphics[width=\linewidth, height=0.65\linewidth,trim= 0 200 466 0,clip]{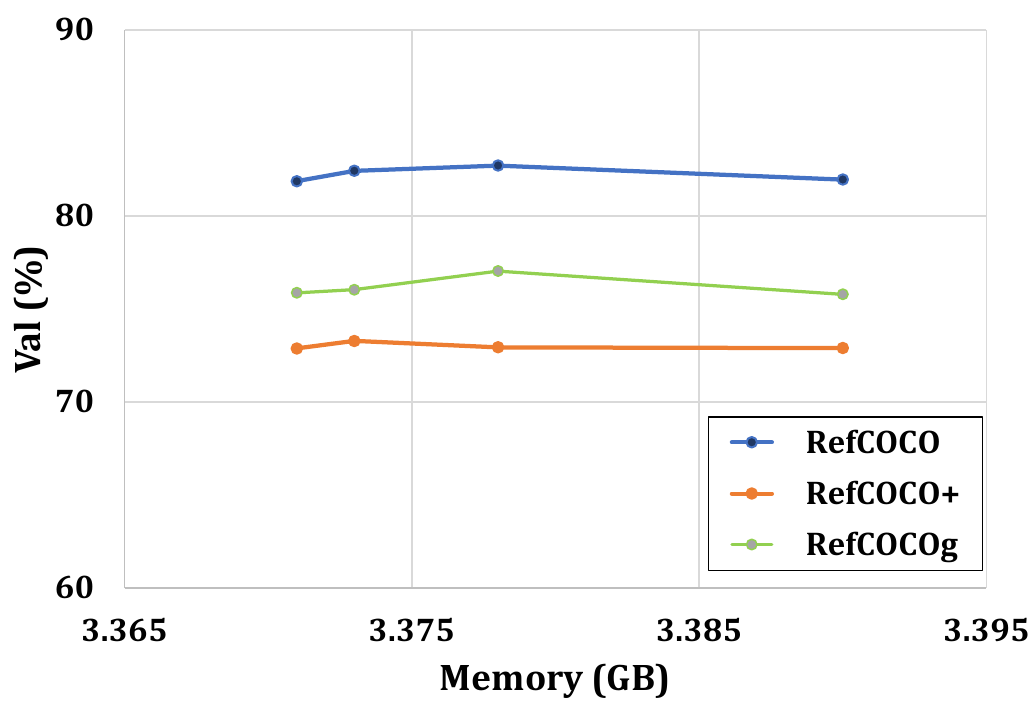}
        \caption{Impact of reduction factor $r\in \{8,4,2,1\}$ on VG.}
        \label{fig:reduction_factor_VG}
    \end{minipage}
\end{figure*}

\subsection{Ablation Studies}

\textbf{Training and Inference Time.}
1) Overall, the inference time of UniPT brings an average of 22.0\% vs. 44.6\% of the best memory-efficient method LST on all VL datasets. 
Actually, the inference time costs are negligible with an extra about 11.7\% consumption on seven tasks including Flickr30K, MSVD, VQA, GQA, RefCOCO, and RefCOCO+/g, which are relatively larger on MSCOCO and MSR-VTT datasets.
2) More importantly, our UniPT simultaneously reduces the training times by a large margin. Specifically, it saves an average of 63.4\% training time of fully fine-tuning. Compared with the counterpart LST, our UniPT is more efficient and resource-friendly.

\textbf{Impact of Reduction Factor $r$.}
In~\Cref{fig:reduction_factor_ITR,fig:reduction_factor_VTR,fig:reduction_factor_QA,fig:reduction_factor_VG}, we adjust the reduction factor
$r\in\{8,4,2,1\}$ and investigate the accuracy-memory trade-off of UniPT. We discover that UniPT is quite robust and stable across a wide range of the reduction factor $r$, which has a better performance and memory trade-off when $r$ is set as 2. 
Note that the Rsum gets better when the reduction factor goes down on the MSVD dataset. 
This may be attributed to the limited size of the training dataset, consisting of only 1200 videos. The overfitting risk tends to escalate as the size of the additional module increases. Unless otherwise stated, the reduction factor $r$ of UniPT is set to 2 across VL tasks in our experiments.

\end{appendix}

\clearpage

{
\small
\bibliographystyle{ieeenat_fullname}
\bibliography{main}
}

\end{document}